\documentclass{article}

\usepackage{PRIMEarxiv}
\usepackage{tabularx}
\usepackage{booktabs}

\usepackage[utf8]{inputenc} 
\usepackage[T1]{fontenc}    
\usepackage{hyperref}       
\usepackage{url}            
\usepackage{booktabs}       
\usepackage{amsfonts}       
\usepackage{nicefrac}       
\usepackage{microtype}      
\usepackage{lipsum}
\usepackage{fancyhdr}       
\usepackage{graphicx}       
\graphicspath{{media/}}     
\usepackage{booktabs}   
\usepackage{tabularx}
\usepackage{pdflscape}
\usepackage{rotating}    
\usepackage{tabularx}  
\usepackage{float}

\usepackage{booktabs,tabularx}
\title{A Survey of Predictive Maintenance Methods: An Analysis of Prognostics via Classification and Regression
}

\author{
  Ainaz Jamshidi \\
  Department of Information Systems\\
  University of Maryland\\
  Baltimore County, USA\\  \texttt{ainazj1@umbc.edu} \\
   \And
  Dongchan Kim \\
  Department of Information Systems \\
  University of Maryland \\
  Baltimore county, USA \\
  \texttt{dkim26@umbc.edu} \\
  \AND
  Muhammad Arif \\
  Department of Information Systems\\
  Colorado State University \\
  Pueblo, USA\\
\texttt{muhammad.arif@csupueblo.edu} \\
}

\begin{document}
\maketitle

\begin{abstract}
Predictive maintenance (PdM) has become a crucial element of modern industrial practice. PdM plays a significant role in operational dependability and cost management by decreasing unforeseen downtime and optimizing asset life cycle management.
Machine learning and deep learning have enabled more precise forecasts of equipment failure and remaining useful life (RUL). Although many studies have been conducted on PdM, there has not yet been a standalone comparative study between regression- and classification-based approaches.
In this review, we look across a range of PdM methodologies, while focusing more strongly on the comparative use of classification and regression methods in prognostics. While regression-based methods typically provide estimates of RUL, classification-based methods present a forecast of the probability of failure across defined time intervals. Through a comprehensive analysis of recent literature, we highlight key advancements, challenges—such as data imbalance and high-dimensional feature spaces—and emerging trends, including hybrid approaches and AI-enabled prognostic systems.
This review aims to provide researchers and practitioners with an awareness of the strengths and compromises of various PdM methods and to help identify future research and build more robust, directed adaptive maintenance systems. Future work may include a systematic review of practical aspects such as public datasets, benchmarking platforms, and open-source tools to support the advancement of PdM research.

\bigskip

\end{abstract}


\begin{table}[ht]
  \centering
  \caption{Nomenclature used in this study.}
  \label{tab:nomenclature}

  \begin{tabularx}{\textwidth}{@{} l X @{}}
    \toprule
    \textbf{Abbreviation} & \textbf{Definition} \\
    \midrule

    PdM      & Predictive Maintenance \\
    RUL      & Remaining Useful Life \\
    ML       & Machine Learning \\
    DL       & Deep Learning \\
    AI       & Artificial Intelligence \\
    IoT      & Internet of Things \\
    CNN      & Convolutional Neural Network \\
    RNN      & Recurrent Neural Network \\
    LSTM     & Long Short‑Term Memory \\
    DI       & Degradation Index \\
    ANN      & Artificial Neural Network \\
    ERNN     & Elman Recurrent Neural Network \\
    SVM      & Support Vector Machine \\
    DES      & Double Exponential Smoothing \\
    MLE      & Maximum Likelihood Estimation \\
    LSM      & Least Squares Method \\
    ARMA     & AutoRegressive Moving Average \\
    MoG‑HMM  & Mixture of Gaussians– Hidden Markov Models \\
    WMA      & Weighted Moving Average \\
    WMS      & Weighted Mean Slope \\

      \bottomrule
  \end{tabularx}
\end{table}

\section{Introduction}
In today's competitive industrial landscape, effective maintenance strategies are essential to ensure operational reliability, cost efficiency, and safety. Equipment downtime caused by unexpected failures not only interrupts production but also significantly escalates operational expenses and jeopardizes overall system productivity and profitability~\cite{jardine2006review}. Maintenance costs are a major part of the total operating costs of all manufacturing or production plants. Depending on the specific industry, maintenance costs can represent between 15 and 60 percent of the cost of goods produced. Maintenance costs billions of dollars spent annually in the US alone~\cite{mobley2002introduction}. Large-scale manufacturing enterprises, in particular, are vulnerable when their equipment fails, leading to significant disruptions. 
\begin{figure}[thp!]
    \centering
    \includegraphics[width=0.75\linewidth]{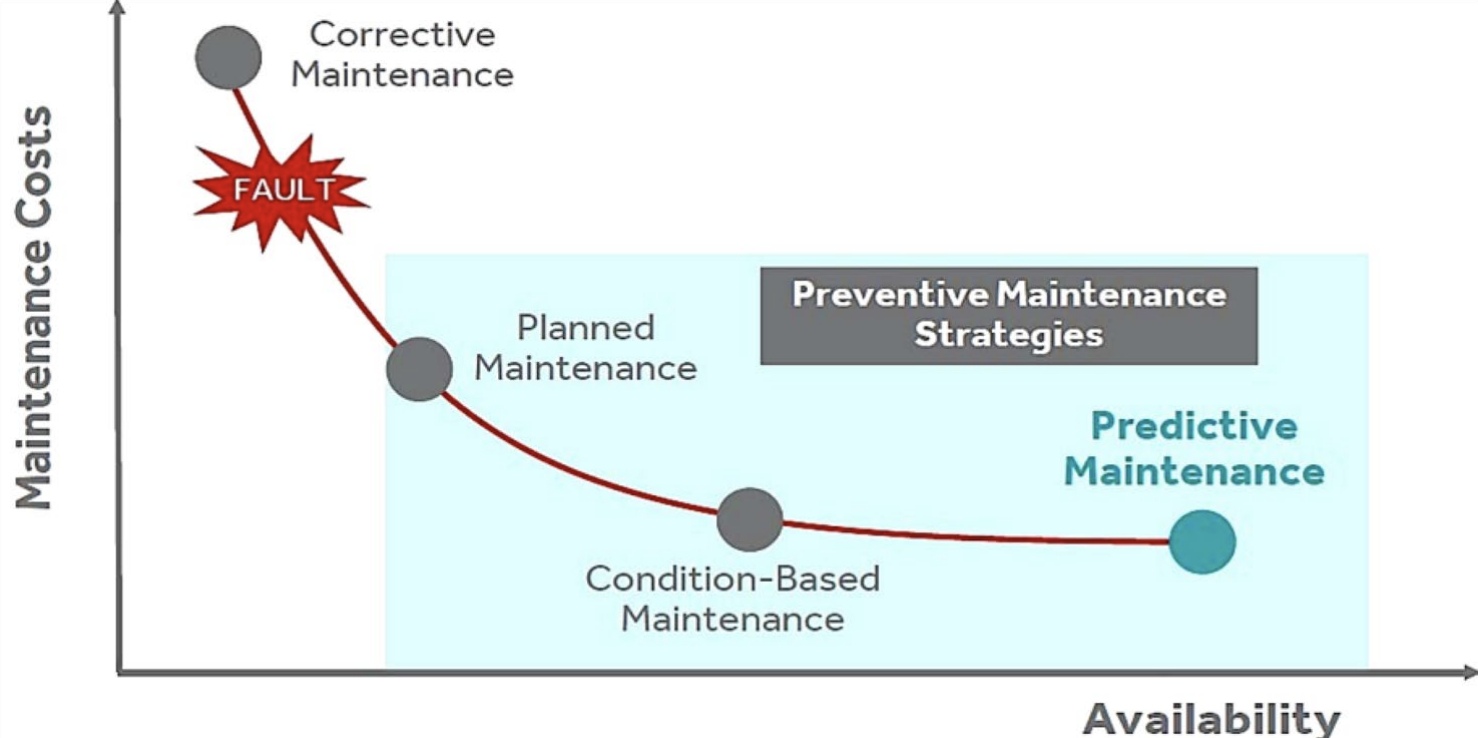}
    \caption{Comparison of maintenance strategies showing how costs decline and equipment availability rises as you move from corrective maintenance (post‐fault) through planned and condition‐based approaches to PdM~\cite{article}.}
    \label{fig:cost}
\end{figure}
Traditional maintenance paradigms, such as corrective and preventive maintenance, have historically dominated industrial practice. As Figure~\ref{fig:cost} shows, overall maintenance cost decreases while equipment availability rises when one moves from purely corrective strategies, through planned and condition-based approaches, to predictive maintenance (PdM)~\cite{article}. 
However, with advancements in sensing technology, data analytics, and machine learning (ML), PdM has emerged as a more sophisticated alternative capable of forecasting failures based on real-time data monitoring and condition-based analysis Unlike preventive maintenance, which often leads to unnecessary maintenance activities. PdM leverages condition-monitoring data and advanced analytical tools to determine optimal maintenance schedules precisely, thereby minimizing both unplanned downtime and excessive maintenance expenditures~\cite{mobley2002introduction, wang2015study}.

Recently, PdM has benefited enormously from recent advances in artificial intelligence (AI) methods. 
Deep‑learning (DL) architectures now dominate RUL forecasting, consistently surpassing classical statistical baselines~\cite{Wu2024,Shah2024}. 
In parallel, probabilistic formulations grounded in survival analysis provide calibrated confidence intervals for each prediction~\cite{Lillelund2024}, while ensemble strategies aggregate heterogeneous learners to boost robustness and generalization~\cite{Wang2025}. 
These AI‑driven techniques deliver highly accurate equipment prognosis—either precise RUL estimates or failure‑probability curves within specified horizons. 
Consequently, PdM has been adopted across diverse sectors: assembly‑line robots and drive‑trains in the automotive industry~\cite{Theissler2021}, wind‑turbine gearboxes and generators in the power‑generation domain~\cite{Shah2024}, and cooling or storage subsystems in high‑performance‑computing (HPC) facilities~\cite{Lima2021}. 
Across these use‑cases, data‑centric PdM improves system reliability, mitigates unplanned outages, and reduces maintenance costs.

This paper systematically surveys the history and current status of PdM methods, particularly data-driven methods and their relative effectiveness as an aspect of industrial prognostics. We focus on regression methods which predict the RUL of equipment, and class-based methods which predict the probability of failure of the equipment in specified future time intervals. In doing so, we hope to identify the best prognostic techniques for meeting the dynamic demands of current-day industrial systems. Nomenclature used in this study are shown in Table~\ref{tab:nomenclature}.

\subsection{Research methodology}

The latest trend in predicting maintenance is, in fact, defined by Industry, where the use of smart technologies such as big data analytics, Internet of Things (IoT) and AI is optimally predictive. With the use of sensor data and machine learning algorithms, predictive maintenance is able to not just predict potential failures but also re-adjust to changing operational conditions providing more flexible maintenance options~\cite{lorenti2023predictive,  biggio2023dynamic}.
Scaife et al. demonstrate that AI enhanced PdM will shorten outages and reduce operational costs by supporting real-time decision-making~\cite{scaife2024improve}. This benefit is useful in sectors such as automotive, power, and high-performance computing, where equipment reliability is crucial for continuous operation~\cite{miller2020system, automotive_2021_review}.

Recently, PdM research includes probabilistic and multi-model deep learning frameworks—including convolutional (CNN) and recurrent neural networks (RNN). These estimates of remaining useful life (RUL), enabling planners to schedule maintenance proactively and keep assets in service longer \cite{deep_learning_2020_survey,AI_reliability_2024}. By moving beyond single-model systems, modern PdM solutions addressed the limitations of traditional methods, particularly in handling complex systems with multiple interconnected components~\cite{multi_model_2021_survey}.

This review discusses the evolution of PdM techniques, focusing on recent advancements in data-driven prognostics, RUL estimation, and condition-based monitoring methodologies. By analyzing current PdM research, particularly in the context of AI and Industry 4.0, this work highlights the potential of PdM to reduce costs, enhance reliability, and align maintenance practices with modern industrial requirements.

\begin{figure}
    \centering
    \includegraphics[width=0.5\linewidth]{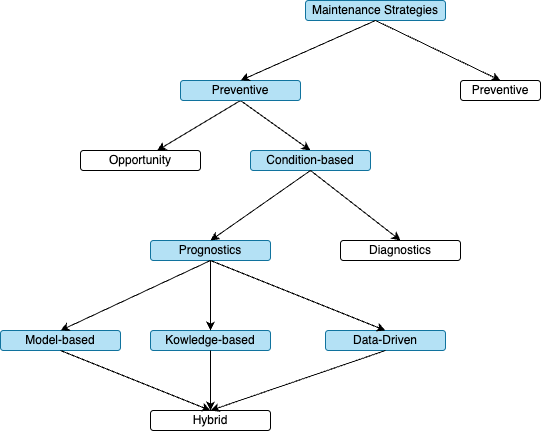}
    \caption{Taxonomy of maintenance strategies, including corrective, preventive, predictive, and hybrid maintenance~\cite{okoh2017predictive}.}
    \label{fig:maintenance_classification}
\end{figure}

\subsection{Contributions}
Our work contributes a structured and in-depth review of PdM techniques, with a focus on comparing regression-based and classification-based approaches. It highlights how regression models are widely used for estimating the RUL of equipment, while classification models are better suited for predicting the likelihood of failure within specific future time windows. This study bridges the gap between academic research and industrial application, and proposes directions for overcoming real-world implementation barriers.

\begin{figure}[h!]
    \centering
    \includegraphics[width=0.95\textwidth]{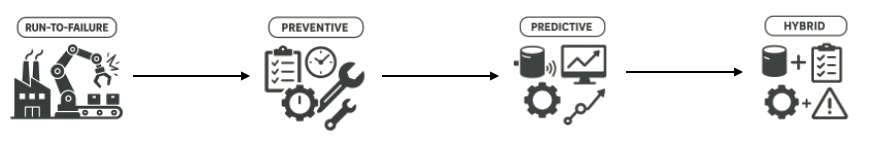}
    \caption{Evolution of major maintenance strategies.}
    \label{fig:maintenance-diagram}
\end{figure}

\section{Evolution of Major Maintenance Strategies}\label{EMMS}
The evolution of maintenance strategies reflects the growing complexity and operational demands of modern equipment. There are four major maintenance strategies as shown in Figures~\ref{fig:maintenance_classification} and~\ref{fig:maintenance-diagram}. Maintenance approaches can be broadly classified into three primary categories: run-to-failure (corrective), preventive, and PdM. In recent years, a fourth category, hybrid maintenance, has emerged, combining elements of these strategies to adapt to specific operational and cost requirements. Summary-level comparison of advantages and disadvantages of each strategy are organized in Table~\ref{table:maintenance_comparison}.

\subsection{Run-to-Failure Maintenance (Corrective Maintenance)}
Corrective maintenance, also known as run-to-failure is the simplest maintenance approach where no actions are taken until a machine fails~\cite{mobley2002introduction,starr2010maintenance}. This strategy is mainly effective for non-critical systems where a sudden change in system will not lead to critical failure. Advantages of this strategy include its simple planning and negligible upfront spending as maintenance is only conducted when absolutely necessary. However, it also can cause major drawbacks, as unexpected failures can lead to unplanned downtime, safety risks, and emergency repair costs, particularly in critical systems~\cite{starr2010maintenance}.
\subsection{Preventive Maintenance (Scheduled Maintenance)}
Preventive maintenance, or scheduled maintenance, 
keep equipment on a predictable service clock, which identifies issues at set intervals rather than checking issues after a failure occurs~\cite{mobley2002introduction,starr2010maintenance}. This approach reduces the unexpected breakdowns by ensuring regular checks and replacements. However, preventive maintenance can result in unnecessary interventions for components that may not require immediate attention. Although this maintenance strategy has drawbacks when an asset’s wear profile is well-characterized, the additional planning burden is generally offset by increased equipment availability and more predictable budgets~\cite{miller2020system}.
\subsection{Predictive Maintenance (Condition-Based Maintenance)}
PdM uses real-time condition monitoring data to forecast equipment failures, thereby allowing for just-in-time maintenance~\cite{mobley2002introduction,selcuk2017predictive}. This approach relies on sensors, machine learning algorithms, and diagnostics to estimate RUL, probabilistic (stochastic) approaches and schedule interventions only when needed based on equipment condition~\cite{deep_learning_2020_survey}. PdM minimizes downtime, extends equipment lifespan, and enhances safety by addressing issues proactively rather than reactively~\cite{lorenti2023predictive,AI_reliability_2024}. PdM's primary components include:
\begin{itemize}
    \item \textbf{Diagnostics:} Focuses on real-time detection and isolation of faults as they arise.
    \item \textbf{Prognostics:} Uses data-driven models to forecast future failures and estimate RUL, enabling proactive maintenance scheduling~\cite{biggio2023dynamic,okoh2017predictive}.
\end{itemize}
PdM is particularly advantageous for industries with high-stakes operational demands, such as power, automotive, and manufacturing sectors~\cite{miller2020system,automotive_2021_review}.

\subsection{Hybrid Maintenance Strategies}
Hybrid maintenance combines aspect of corrective, preventive, and predictive practices to create a flexible, efficient framework tailored to the needs of complex systems~\cite{automotive_2021_review,lima2021smart}. By pairing predictive insights with routine inspections and readiness for corrective action, organizations can plan maintenance precisely while remaining responsive to unexpected issues. This balance improves adaptability in environments where operational demands and equipment health vary significantly~\cite{AI_reliability_2024,multi_model_2021_survey}.

Hybrid maintenance is particularly valuable in high performance settings such as high-performance computers and automotive sectors where every downtime is essential~\cite{multi_model_2021_survey}. Many hybrid systems use multi-model and ensemble approaches, employing machine learning algorithms that can assess various aspects of equipment health and improve the accuracy of maintenance predictions~\cite{biggio2023dynamic}. This strategy would offer a cost-effective and reliable solution to balance downtime risk, maintenance cost, and equipment reliability in dynamic and complex system.
\begin{figure}[htp!]
    \centering
    \includegraphics[width=0.95\linewidth]{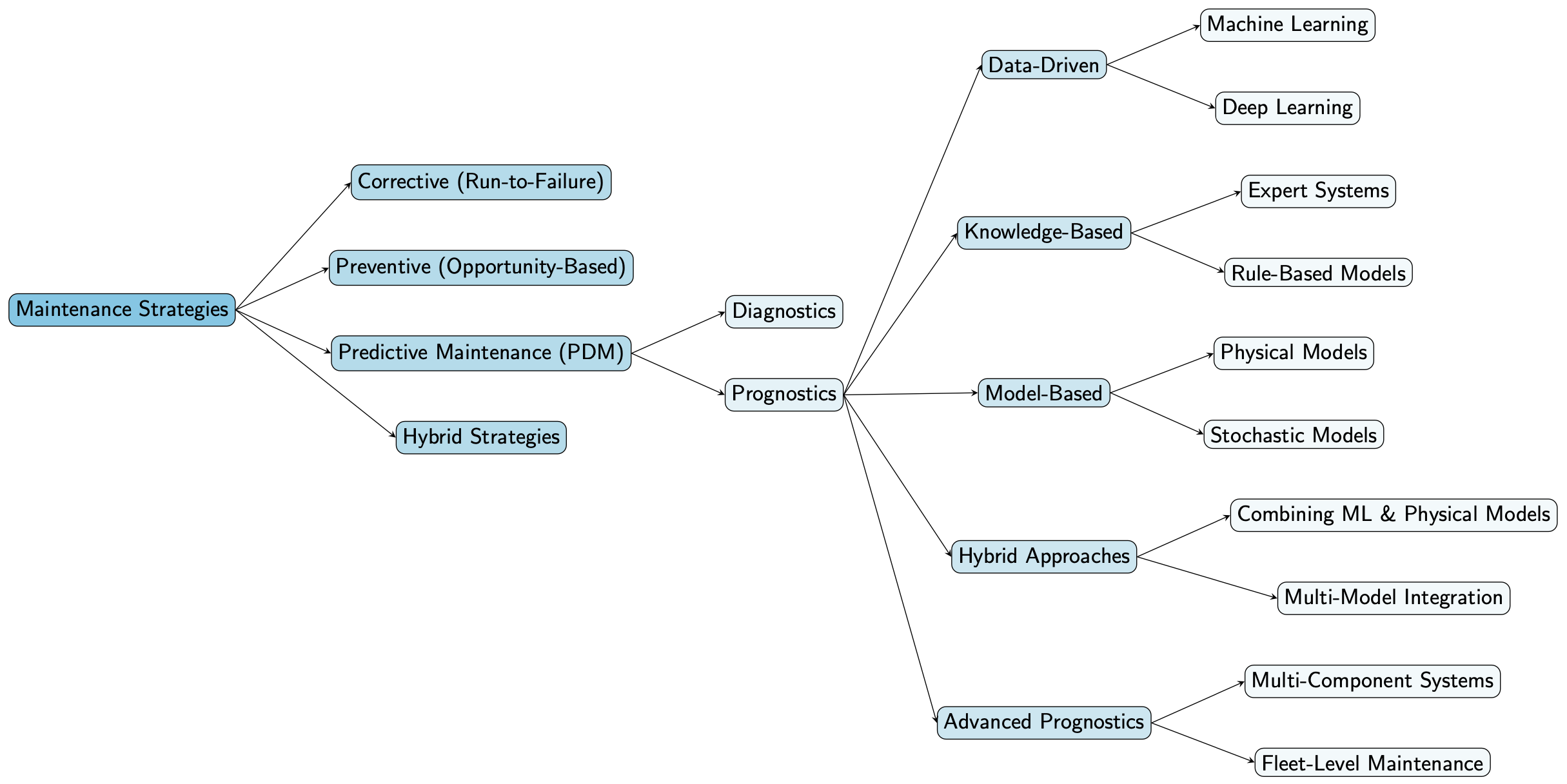}
    \caption{An overview of maintenance strategies structured around the prognostics taxonomy, highlighting method-level approaches.}
    \label{fig:wideTax}
\end{figure}

\newcolumntype{Y}{>{\raggedright\arraybackslash}X}

\begin{table}[ht]
  \centering
  \caption{Comparison of Major Maintenance Strategies.}
  \label{table:maintenance_comparison}

  \begin{tabularx}{\textwidth}{@{} l Y Y Y Y @{}}
    \toprule
    \textbf{Strategy} & \textbf{Approach} & \textbf{Advantages} &
    \textbf{Disadvantages} & \textbf{Key Ref.}\\
    \midrule                

    Run-to-Failure (Corrective) &
      No maintenance until equipment fails &
      • Simplest planning\newline • Minimal upfront cost &
      • Unexpected downtime\newline • Costly repairs\newline • Safety risks &
     ~\cite{mobley2002introduction,starr2010maintenance} \\
    \midrule                

    Preventive (Scheduled) &
      Regular service at fixed intervals &
      • Fewer sudden failures\newline • Predictable budgeting &
      • Possible over-maintenance\newline • Higher planning overhead &
     ~\cite{miller2020system,starr2010maintenance} \\
    \midrule

    Predictive (Condition-Based) &
      Data-driven service triggered by condition indicators &
      • Minimises unplanned stops\newline • Extends asset life\newline • Improves safety &
      • High sensor / IT cost\newline • Requires analytics expertise &
     ~\cite{AI_reliability_2024,selcuk2017predictive} \\
    \midrule

    Hybrid &
      Mix of corrective, preventive, and predictive tactics &
      • Flexible\newline • Tailored to context\newline • Potentially most efficient &
      • Implementation complexity\newline • Advanced analytics needed &
     ~\cite{multi_model_2021_survey,lima2021smart} \\
    \bottomrule
  \end{tabularx}
\end{table}

\begin{figure}[htp!]
    \centering
    \includegraphics[width=0.8\linewidth]{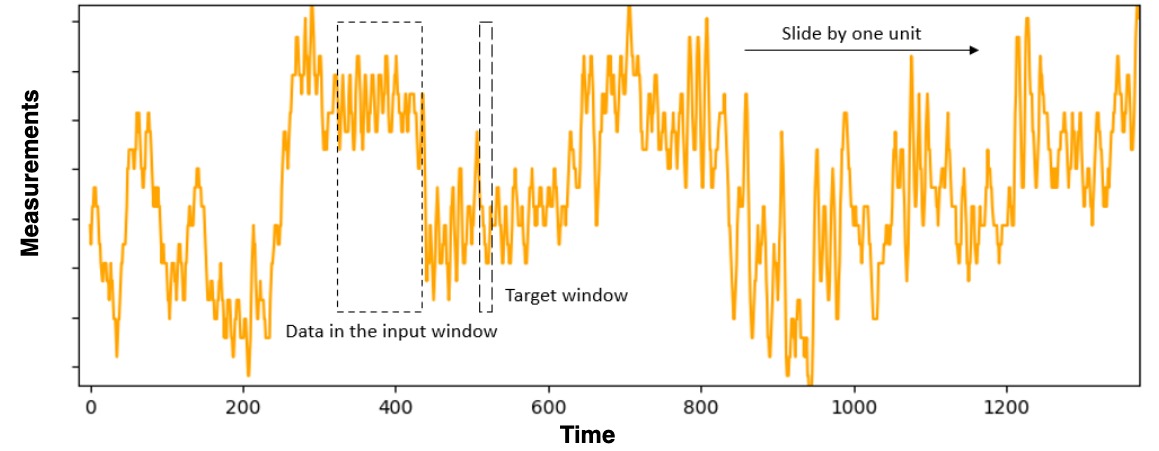}
    \caption{A depiction of the sliding window method to prepare input time series data for
machine learning and deep learning models.}
    \label{fig:SlidingW}
\end{figure}

\section{Prognostic Approaches}\label{PA}
Prognostic approaches aim to detect early indicators of failure and predict the remaining useful life before a breakdown occurs. Generally, machines experience degradation prior to failure; thus, monitoring degradation trends can help identify incipient faults and enable timely corrective actions to prevent failures~\cite{yan2004prognostic}. Prognostic approaches aim to detect early indicators of failure and
predict the RUL before a breakdown occurs.
Figure~\ref{fig:SlidingW} illustrates how a sliding-window technique segments the raw time series signal into overlapping subsequences; these windows become well-structured inputs for ML and DL models that learn the degradation trajectory and predict the RUL. The International Standard Organization (ISO 13381-1:2004) defines prognostics as “the estimated-time-to-failure and the risk of existence or subsequent appearance of one or more failure modes”~\cite{medjaher2012remaining}.

Prognostic models are commonly classified into four categories: data-driven, experimental (knowledge-based), physical (model-based), and hybrid approaches~\cite{peng2010current}. A method-level overview of prognostics approaches, proposed in existing literature, is shown in Figure~\ref{fig:wideTax}. 
In this survey, we review previous studies that focus on the data-driven approach. A concise summary of representative data-driven prognostic approaches is provided in Table~\ref{tab:merged_rul_methods}.
Prognostic approach—also known as the machine learning or data mining approach~\cite{schwabacher2005survey}—employs algorithms that learn predictive models directly from data.
Broadly, prognostics data-driven strategy includes several main steps. The first step is data acquisition which includes process of collecting and storing information for the purpose of maintenance. The acquired data can be divided into two main categories: event data and condition monitoring data. Event data includes information on equipment-related occurrences such as breakdowns and maintenance activities, while condition monitoring data is more diverse, encompassing vibration measurements, acoustic signals, oil analysis, temperature, pressure, moisture, humidity, and environmental or weather-related parameters~\cite{jardine2006review,okoh2017predictive}.

Model-based strategy includes physical or hybrid strategies. 
In Wang’s bearing-life model~\cite{wang2002model}, Wang et al. (2002) coupled a two-stage delay-time failure process with a Bayesian stochastic filter.
Each incoming vibration sample refines the posterior distribution of the remaining useful life, and a cost-rate function explicitly balances preventive- and run-to-failure expenses. Wang's study illustrates how model-based degradation dynamics and probabilistic updating can be integrated into a practical maintenance-decision loop.

Historical data can be used to estimate parameters and automatically learn a model of the system’s behavior. Following data acquisition, the next step is parameter estimation, which can be performed using methods such as least squares (LSM), maximum likelihood estimation (MLE), various machine learning techniques, and Bayesian parameter estimation methods. After parameter estimation, prognostics modeling and decision making follow, where the estimated parameters are incorporated into the model. The output of the model then plays a pivotal role in informing maintenance decisions.

There are two main prediction types in machine prognostics. The first—and most widely used—is to predict the RUL, which is the time left before a failure occurs, based on the current condition and past operational profile of the machine. The second type estimates the probability that a machine will operate without failure over a specified future period, given the same information. Although only a few studies have addressed this second type, further research is needed to explore its full potential.

\begin{sidewaystable}
\centering
\caption{Summary of the various prediction approaches presented by different authors.}
\label{tab:merged_rul_methods}
\begin{tabularx}{\linewidth}{|c|X|X|X|X|X|}
\hline
\textbf{Reference} & \textbf{Input Type} & \textbf{Output Type} & \textbf{ML/DL Approach} & \textbf{Transformation} & \textbf{Feature Selection / Extraction} \\
\hline

\tiny{Asmai et al. (2014)}~\cite{asmai2014time} & \tiny{Time series + Transformation (logistic)} & \tiny{RUL} & \tiny{ANN and Double Exponential Smoothing (DES)} & \tiny{-} & \tiny{Yes} \\
\hline
\tiny{Ahmadzadeh et al. (2013)}~\cite{ahmadzadeh2013remaining} & \tiny{Time series} & \tiny{RUL} & \tiny{ANN} & \tiny{-} & \tiny{Yes (PCA)} \\
\hline
\tiny{Bey et al. (2009)}~\cite{bey2009practical} & \tiny{Time series} & \tiny{RUL} & \tiny{Decision tree (classifier), weighted mean slope (WMS)} & \tiny{-} & \tiny{Yes (PCA)} \\
\hline
\tiny{Yan et al. (2004)}~\cite{yan2004prognostic} & \tiny{Time series + Transformation (logistic)} & \tiny{RUL} & \tiny{Logistic regression, ARMA} & \tiny{Logistic transformation} & \tiny{No} \\
\hline
\tiny{Hota et al. (2017)}~\cite{hota2017time} & \tiny{Time series + Transformation (normalization)} & \tiny{Next Value Prediction of Stock Price} & \tiny{Sliding Window Based RBF NN} & \tiny{Normalization} & \tiny{Yes} \\
\hline
\tiny{Goebel et al. (2008)}~\cite{goebel2008comparison} & \tiny{Time series (log space transformation)} & \tiny{RUL} & \tiny{RVM, GPR, Neural Network} & \tiny{Logarithmic transformation} & \tiny{No} \\
\hline
\tiny{Tobon-Mejia et al. (2012)}~\cite{tobon2012data} & \tiny{Time series + WPD} & \tiny{RUL} & \tiny{Mixture of Gaussians, MoG-HMM} & \tiny{WPD} & \tiny{Yes} \\
\hline
\tiny{Louen et al. (2013)}~\cite{louen2013new} & \tiny{Time series + Transformation} & \tiny{RUL} & \tiny{SVM, Weibull reliability function} & \tiny{Transformation} & \tiny{Yes} \\
\hline
\tiny{Le et al. (2014)}~\cite{le2014predictive} & \tiny{Time series} & \tiny{RUL} & \tiny{Linear regression, Gaussian kernel method} & \tiny{-} & \tiny{Yes (t-test)} \\
\hline
\tiny{Susto et al. (2012)}~\cite{susto2012predictive} & \tiny{Time series} & \tiny{RUL} & \tiny{Gaussian kernel estimation, Kalman predictor} & \tiny{-} & \tiny{Yes (temp-based)} \\
\hline
\tiny{Tran et al. (2012)}~\cite{pham2012machine} & \tiny{Time series} & \tiny{RUL} & \tiny{Proportional hazard model, SVM} & \tiny{-} & \tiny{Yes} \\
\hline
\tiny{Yu et al. (2006)}~\cite{yu2006feature} & \tiny{Time series} & \tiny{Next feature value prediction} & \tiny{Elman recurrent neural network (ERNN)} & \tiny{-} & \tiny{Yes (PCA)} \\
\hline
\tiny{Okoh et al. (2017)}~\cite{okoh2017predictive} & \tiny{Time series} & \tiny{RUL} & \tiny{Weibull function} & \tiny{-} & \tiny{No} \\
\hline


\tiny{Wang et al. (2002)}~\cite{wang2002model} & \tiny{Time series} & \tiny{RUL} & \tiny{Stochastic filtering theory} & \tiny{-} & \tiny{Yes} \\
\hline
\tiny{Liao et al. (2005)}~\cite{liao2005predictive} & \tiny{Time series + Nature Log Transformation} & \tiny{RUL} & \tiny{Proportional hazards model} & \tiny{Logarithmic transformation} & \tiny{Yes} \\
\hline
\tiny{Susto et al. (2016)}~\cite{susto2016dealing} & \tiny{Time series} & \tiny{RUL} & \tiny{Ridge regression module} & \tiny{-} & \tiny{Yes (Supervised Aggregative Feature Extraction)} \\
\hline
\tiny{Li et al. (2018)} ~\cite{li2018remaining}& \tiny{Time series + Normalization} & \tiny{RUL} & \tiny{DCNN + Dropout} & \tiny{Normalization} & \tiny{Feature selection} \\
\hline
\tiny{Ren et al. (2017)}~\cite{ren2017multi} & \tiny{Time series + Normalization} & \tiny{RUL} & \tiny{Deep Neural Network} & \tiny{Normalization} & \tiny{Time and frequency features} \\
\hline
\tiny{Babu et al. (2016)}~\cite{babu2016deep} & \tiny{Time series + Normalization} & \tiny{RUL} & \tiny{Deep CNN} & \tiny{Normalization} & \tiny{Yes (via CNN)} \\
\hline
\tiny{Zhang et al. (2017)}~\cite{zheng2017long} & \tiny{Time series} & \tiny{RUL} & \tiny{MODBNE} & \tiny{-} & \tiny{Feature extraction} \\
\hline
\tiny{Zheng et al. (2017)}~\cite{zheng2017long} & \tiny{Time series + Normalization} & \tiny{RUL} & \tiny{LSTM} & \tiny{Normalization} & \tiny{-} \\
\hline
\tiny{Yuan et al. (2016)}~\cite{yuan2016fault} & \tiny{Time series} & \tiny{RUL} & \tiny{RNN, GRU-LSTM, Vanilla LSTM} & \tiny{-} & \tiny{-} \\
\hline
\tiny{Fulp et al. (2008)}~\cite{fulp2008predicting} & \tiny{Time series + Transformation} & \tiny{Time window} & \tiny{Spectrum-kernel SVM} & \tiny{Transformation} & \tiny{Yes} \\
\hline
\tiny{Martinez et al. (2017)}~\cite{martinezsequence} & \tiny{Time series + Transformation} & \tiny{RUL + Classification (time window prediction)} & \tiny{Random Forest, GMM, LSTM, RNN} & \tiny{Transformation} & \tiny{Yes (High cost repairs)} \\
\hline
\tiny{Prytz et al. (2018)}~\cite{prytz2013analysis} & \tiny{Time series} & \tiny{Time window} & \tiny{KNN, C5.0, Random Forest} & \tiny{-} & \tiny{Yes (expert selected)} \\
\hline
\tiny{Satta et al. (2017)}~\cite{satta2017dissimilarity} & \tiny{Time series} & \tiny{Time window} & \tiny{Adaboost, Decision Tree} & \tiny{Dissimilarity-based approach} & \tiny{Yes (paper novelty)} \\
\hline
\tiny{Susto et al. (2015)}~\cite{susto2015machine} & \tiny{Time series (Tool/logistic variables)} & \tiny{Time window (Fault Horizon)} & \tiny{SVM + KNN} & \tiny{-} & \tiny{Yes} \\
\hline
\tiny{Battifarano et al. (2018)}~\cite{battifarano2018predicting} & \tiny{Time series} & \tiny{Time window (24 hours)} & \tiny{Logistic Regression} & \tiny{-} & \tiny{Both} \\
\hline

\end{tabularx}
\end{sidewaystable}

\subsection{Regression-based PDM}
\begin{figure}
    \centering
    \includegraphics[width=0.85\linewidth]{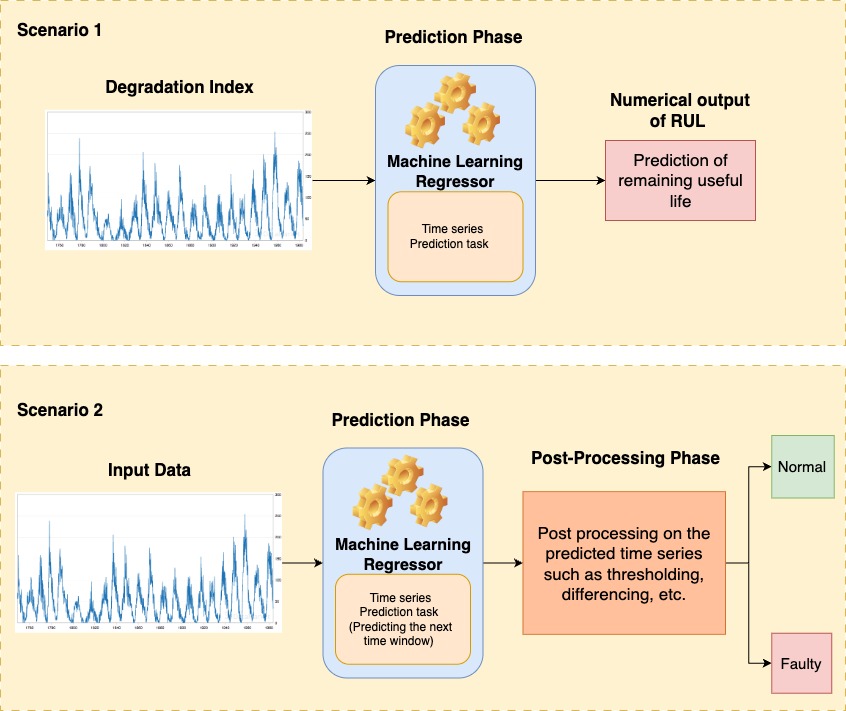}
    \caption{Overview of two regression-based predictive‐maintenance workflows: (1) direct regression of a degradation index time series to predict Remaining Useful Life, and (2) short‐term time‐series forecasting on the input data followed by threshold‐based post-processing to classify the future time frame as Normal or Faulty.}
    \label{fig:reg_method}
\end{figure}
In PdM problems, regression-based formulations are commonly used to predict the RUL of equipment~\cite{Khan2022}. RUL can be defined as the difference between the end of useful life and the current time~\cite{louen2013new}. Broadly speaking, there are two main regression-based PdM workflows (Figure~\ref{fig:reg_method}). The first is direct RUL regression, which trains a model on a degradation index to directly estimate how much useful time remains before a component fails. The second approach is time series forecasting, which involves training a model on historical sensor data to predict future values. In this case, faults are identified by applying predefined thresholds to the forecasted data, allowing the system to flag when a component is likely to degrade beyond acceptable limits. 

Numerous studies have explored regression models for RUL prediction.
Ahmadzadeh et al. (2013) used an artificial neural network (ANN) with PCA-based feature reduction to estimate the RUL of grinding mill liners~\cite{ahmadzadeh2013remaining}. Asmai et al. (2014) combined ANN and Double Exponential Smoothing to capture both linear and nonlinear patterns in cutting tool degradation data~\cite{asmai2014time}. Similarly, Yu et al. (2006) utilized an Elman Recurrent Neural Network to forecast tool behavior over its lifecycle~\cite{yu2006feature}. Hota et al. (2017) used a sliding window and Weighted Moving Average (WMA) for preprocessing time series data, followed by a Radial Basis Function Network (RBFN) to predict future values~\cite{hota2017time}.

Several other regression techniques have been compared in challenging environments. Goebel et al. (2008) evaluated relevance vector machines (RVM), Gaussian process regression (GPR), and neural networks under sparse, noisy conditions~\cite{goebel2008comparison}. Le et al. (2014) combined linear regression with a Gaussian kernel method to estimate the prediction error distribution for etching chamber diagnostics~\cite{le2014predictive}. Louen et al. (2013) introduced a two-step framework using an support vector machine (SVM) classifier followed by a Weibull reliability function for RUL estimation~\cite{louen2013new}.

Other hybrid or statistical models include decision tree classifiers with weighted mean slope (WMS) post-processing~\cite{bey2009practical}, proportional hazards models~\cite{liao2005predictive}, and kernel-based probabilistic filters such as Kalman or particle filters~\cite{susto2012predictive}. Tran et al. (2012) presented a three-stage method for assessing the health of machine condition and forecasting the remaining useful life~\cite{pham2012machine}. Tobon-Mejia et al. (2012) integrated wavelet packet decomposition and MoG-HMMs for complex failure modeling~\cite{tobon2012data}. Logistic regression and ARMA were jointly applied by Yan et al. (2004) to estimate elevator door system lifetimes~\cite{yan2004prognostic}.

Deep learning methods have gained popularity for their superior ability to learn from raw sensor data. Li et al. (2018) introduced a deep convolutional neural network (CNN) that outperformed other deep models such as RNN, LSTM, and DNN in RUL prediction~\cite{li2018remaining}. Babu et al. (2016) proposed a CNN-based regressor trained on multivariate time series data~\cite{babu2016deep}. Zheng et al. (2017) demonstrated that LSTM networks outperform CNNs and traditional models, particularly in capturing temporal degradation patterns under varying operating conditions~\cite{zheng2017long}. Yuan et al. (2016) confirmed these results by comparing multiple LSTM architectures~\cite{yuan2016fault}.

More advanced architectures include Zhang et al.’s (2017) Multiobjective Deep Belief Network Ensemble (MODBNE), which uses evolutionary algorithms to build diverse ensembles for robust predictions~\cite{zhang2016multiobjective}. Ren et al. (2017) fused time-domain and frequency-domain features using a deep neural network to estimate RUL of bearings~\cite{ren2017multi}. Lastly, Okoh et al. (2017) developed a statistical modeling framework using Weibull CDFs to model synthetic failure data and inform lifecycle decisions in aerospace maintenance~\cite{okoh2017predictive}.

These approaches illustrate the diversity and evolution of regression-based prognostic techniques, with deep learning increasingly favored for complex systems and high-dimensional sensor data, as also emphasized in recent literature surveys~\cite{Wu2024,adryan2025predictive}.

\subsubsection{Summary and Discussion}

This section outlines two primary approaches discussed in the literature for predicting remaining useful life. The first approach is calculating the value of RUL directly via feeding the data into a machine learning algorithm~\cite{ahmadzadeh2013remaining} or a time series prediction technique~\cite{okoh2017predictive}. The second approach exploits a hierarchical method to estimate RUL which its first step is generating degradation index (DI). DI can be the performance or health indication of equipment. In order to generate DI, Most studies employed logistic regression to transfer the raw condition monitoring data to the failure probabilities. In the second step, the time series prediction techniques are used in the prognostics model to extrapolate the series of degradation value into the future time~\cite{asmai2014time,louen2013new}. Using failure probabilities can be advantageous since there are many types of equipment in the industry that can produce various operating condition monitoring data. Hence, DI can be used as a generalized input for prognostic models regardless of the different types of condition monitoring data and quantity~\cite{asmai2014time}; in other words, using methods that maps the high dimensional feature space to the lower one leads to computational cost reduction.

\subsection{Classification-based PDM}
\begin{figure}[htp!]
    \centering
    \includegraphics[width=0.8\linewidth]{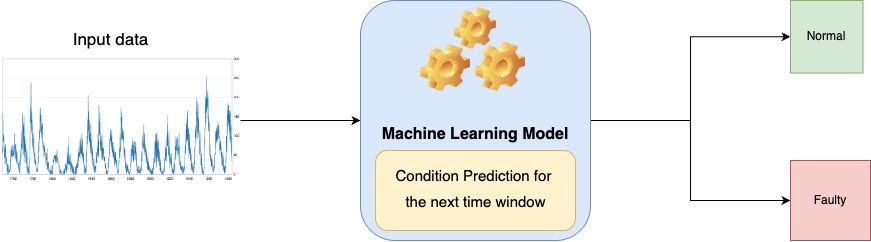}
    \caption{High-level overview of utilizing machine learning in a predictive maintenance pipeline.}
    \label{fig:ML_class}
\end{figure}

In this section, we aim to survey studies that aimed to predict the time window in which failures would occur using ML as shown in Figure~\ref{fig:ML_class}. Fulp et al. (2008) proposed a new spectrum kernel SVM to predict failure events~\cite{fulp2008predicting}. This approach was evaluated experimentally using actual system log files to predict hard disk failures. Results indicated the introduced method can predict hard drive failure with 73\% accuracy two days in advance. Martinez et al. (2017) used a combination of supervised classification and RUL to predict time to failure (predict future repairs) along with the failure type of a commercial vehicles company~\cite{martinezsequence}. Models employed for this purpose are random forest (RF), gaussian mixture model, LSTM Recurrent Neural Networks. Prytz et al. (2018) employed and compared three different classifiers; K-nearest neighbor(KNN), C5.0, and RF, to foretell failures in a prediction horizon~\cite{prytz2013analysis}. The result of this investigation revealed that RF is a better classifier. Satta et al. (2017) proposed a novel dissimilarity-based approach to select meaningful features~\cite{satta2017dissimilarity}. Their adopted classifier (decision tree with AdaBoost algorithm) has been trained with the selected feature by proposed method to predict a time window where the failure is likely to happen. Susto et al. (2015) proposed a multiple classifiers approach to predict occurring fault horizon. They took advantage of SVM and KNN classifiers in their study~\cite{susto2015machine}. 

\subsubsection{Summary and Discussion}

Two data-centric obstacles of PdM research consist of class imbalanced and high dimensionality. A dataset is considered imbalance when one class contains far more samples than the other class, causing conventional learning algorithms to favor the majority class and deliver poor minority class recall. Mitigation strategies include using more informative evaluation metrics, applying re-sampling technique and training cost-sensitive models~\cite{kubat1997addressing}. 

The second obstacle is called the curse of dimensionality: datasets with thousands of features, often exceeding the number of observations, inflate computational cost and erode model generalization. Dimensionality reduction techniques such as LDA and PCA are commonly used to resolve these problems. The following section reviews recent studies that confront these two challenges. Prytz et al. (2013) studied an imbalanced, high‑dimensional dataset and introduced a cost‑saving objective that depends on true positives, false positives, and the relative costs of planned versus unplanned breakdowns~\cite{prytz2013analysis}. Because simple accuracy is misleading under severe imbalance, they reported the popular F‑score instead. Among the classifiers evaluated—KNN, RF, and C5.0—RF and C5.0 performed best, with RF giving the larger net cost savings owing to a higher surplus of true positives over false positives. Martínez et al. (2017) tackled a similarly challenging dataset and used average recall to compensate for imbalance. They compared HMM, LSTM, GMM, and RF models, finding RF again achieved the highest average recall~\cite{martinezsequence}.
Susto et al. (2015) proposed a multiple‑classifier scheme in which several models run in parallel~\cite{susto2015machine}. A maintenance action is triggered by the classifier that minimizes an explicit cost function. Tested on six features, their framework showed that SVM outperformed KNN on both cost and predictive accuracy. Battifarano et al. (2018) applied logistic regression to an imbalanced dataset containing 30 attributes per record. They weighted failure samples 100:1 relative to non‑failures, enabling the model to perfectly detect all failures and almost perfectly classify non‑failures~\cite{battifarano2018predicting}. Ahmadzadeh et al. (2014) used an ANN to estimate remaining useful life. Although their dataset had only nine features, the ANN captured the nonlinear input–output relations and achieved roughly 90\% accuracy~\cite{ahmadzadeh2014remaining}.

\section{Conclusion}
This survey reviewed the rapidly evolving world of PdM methodologies, emphasizing classification-based and regression-based prognostics methods. With the growing use of data-centered maintenance practices in industries, the ability to predict failures and predict RUL has never been more important. This review covered a broad array of models, including traditional machine learning processes and deep-learned structures, and highlighted their relative strengths and limitations, and how relevant they would be for different industrial situations.
While regression processes generally outperform better for RUL predictions based on direct interpretability and easily output RUL; classification-based processes provide valuable insights when the goal is to predict failure to occur within a defined time horizon. Each paradigm offers unique advantages, and results will clearly be governed by a variety of factors including data, sensor resolution, and operating conditions.
There have been improvements, but many obstacles remain--especially regarding issues of data imbalance, high dimensionality, and the ability to practically integrate one of the models in real-world systems. Future research could lead to standardized datasets, higher interpretability of models, and more hybrid models with regression and classification.
Ultimately, the evolution of PdM model is going to depend on balancing algorithmic advancement and real-world relevance, ensuring maintenance decisions are algorithmically valid and operationally realistic.

Moving forward, an important avenue is to systematically survey strategies for coping with the high-dimensional, imbalanced, and often noisy character of PdM data—reviewing methods that handle sparsity and missing values while presenting comparative case studies of preprocessing techniques such as PCA, autoencoders, and data-augmentation pipelines.

\bibliographystyle{unsrt}  
\bibliography{Refs}

\end{document}